\documentclass{article}

\title{Faster Symmetry Breaking Constraints for Abstract Structures}

\author{
    {\"O}zg{\"u}r Akg{\"u}n, 
    {Mun See} Chang, 
    Ian P.\ Gent, 
    Christopher Jefferson
}

\usepackage{xspace}
\usepackage{amsmath}
\usepackage{amssymb}
\usepackage{mathabx}
\usepackage{url}

\usepackage{amsthm} 
\newtheorem{lemma}{Lemma}[section]

\theoremstyle{definition}
\newtheorem{definition}{Definition} [section]
\newtheorem{example}{Example} [section]

\theoremstyle{remark}
\newtheorem{remark}{Remark}[section]

\newcommand{\essence}[0]{\textsc{Essence\xspace}\xspace}
\newcommand{\essencePrime}[0]{\textsc{Essence Prime\xspace}\xspace}
\newcommand{\conjure}[0]{\textsc{Conjure\xspace}\xspace}

\newcommand{\savilerow}[0]{\textsc{SavileRow}\xspace}
\newcommand{\minion}[0]{\textsc{Minion}\xspace}
\newcommand*{\dotleq}{\mathrel{\dot{\leq}}}

\newcommand*{\symleq}{\mathrel{\preceq}}
\newcommand*{\symless}{\mathrel{\prec}}
\newcommand{\Dom}[0]{\mathrm{Dom}}
\newcommand{\Val}[0]{\mathrm{Val}}
\newcommand{\Sym}[0]{\mathrm{Sym}}

\newcommand{\lastPaper}[0]{\cite{unnamedTypeSB}\xspace}

\usepackage{cleveref}
\usepackage{appendix}
\AtBeginEnvironment{appendices}{\crefalias{section}{appendix}}

\begin{document}

\maketitle

\begin{abstract}
In constraint programming and related paradigms, a modeller specifies their problem in a modelling language for a solver to search and return its solution(s).
Using high-level modelling languages such as \essence, a modeller may express their problems in terms of \emph{abstract} structures. 
These are structures not natively supported by the solvers, and so they have to be transformed into or \emph{represented} as other structures before solving. 
For example, nested sets are abstract structures, and they can be represented as matrices in constraint solvers. 
Many problems contain symmetries and one very common and highly successful technique used in constraint programming is to ``break'' symmetries, to avoid searching for symmetric solutions. This can speed up the solving process by many orders of magnitude.
Most of these symmetry-breaking techniques involve placing some kind of ordering for the variables of the problem, and picking a particular member under the symmetries, usually the smallest.
Unfortunately, applying this technique to abstract variables produces a very large number of complex constraints that perform poorly in practice.
In this paper, we demonstrate a new incomplete method of breaking the symmetries of abstract structures by better exploiting their representations. 
We apply the method in breaking the symmetries arising from indistinguishable objects, a commonly occurring type of symmetry, and show that our method is faster than the previous methods proposed in \lastPaper. 
\end{abstract}

\section{Introduction}
\label{sec: intro}

High-level constraint modelling languages such as \essence \cite{akgun2022conjure} allow users to state their problems (or models) in a higher level of abstraction. 
For example, a user may specify their problems in terms of combinatorial structures such as partitions, relations and multisets. 
As constraint solvers cannot solve problems with such abstract objects directly, automatic model rewriting tools \conjure and \savilerow can rewrite such a problem specification into equivalent ones suitable for a solver. 
Through this automated rewriting system, a domain expert no longer needs to be burdened with the task of writing long, error-prone, low-level models, and choosing among their equivalents the most performant model for the selected solver. 
Therefore, the solving technologies can now be made more widely available.

It is well known that the presence of symmetries can significantly impede solver performance, as the solver would spend time exploring equivalent parts of the search space. Therefore, there has been a significant amount of research on how to remove symmetries in a process called \emph{symmetry breaking} -- see \cite{GENT2006329} for an overview. 
The most widely used method of symmetry breaking is lex-leader symmetry breaking constraints, which force assignments considered during search to be the smallest value among their symmetric equivalences.

The lex-leader constraints for a set of symmetries $G$ are the set of constraints $X \leq X^g$ for every symmetry $g \in G$, where $X$ is the list of all variables, and $X^g$ is the result of applying the symmetry $g$ to $X$. In practice, the size of the set of symmetries $G$ is too large to add all such constraints, so we instead add constraints for some subset of $G$. There are a number of ways of choosing this set \cite{10.1007/978-3-642-23786-7_55}, the most notable of which is  ``double-lex'' \cite{rowColSym}.

While we need to limit the number of constraints we generate, the individual constraints $X \leq X^g$ are usually expressed quite compactly.
For many simple cases of symmetries, such as permutations of the elements of a list, each lex-leader constraint can be expressed as a single lexicographic ordering constraint. Other simple symmetries, such as ``value symmetries'' (which permute the values of variables without permuting the values of different variables), can also be expressed compactly and efficiently \cite{Law2006}. However, when we have more complex and abstract types, such as a set of sets of integers, requiring a set to be smaller than only one symmetric equivalent still requires a large set of complex constraints. 
We will demonstrate in our experiments that this can slow down the solving process significantly.

In this paper, we propose a new symmetry handling technique for abstract structures, which we call ``delayed symmetry application''. This method has the advantage that it always produces a single lexicographic ordering constraint for each lex-leader constraint on abstract variables. 
While this constraint breaks less symmetries than the traditional lex-leader constraints, we will show that in practice it greatly improves performance, allowing us to perform efficient symmetry breaking on a wide range of high-level types.

We apply our method to breaking the symmetries induced by \emph{unnamed types}. Unnamed types are introduced into the high-level modelling language \essence to encapsulate indistinguishable objects, one of the most common sources of symmetries. 
\lastPaper proposed methods of breaking the symmetries arising from unnamed types using the lex-leader constraints. 
As an \essence structure can be abstract, we show that our symmetry breaking is better when we have models with abstract structures. 

We begin the paper with the background on symmetry breaking in constraint programming and high-level modelling, as well as a summary of \lastPaper. 
Our method requires a new ordering of abstract objects, which we shall define in \Cref{section: dynamic ordering}. 
We then give our method in \Cref{sec: fast sym break}.
We see how the method is useful for breaking the symmetries of unnamed types in \Cref{sec: application}.

\section{Background}
\label{sec:background}

A \emph{constraint satisfaction problem} (CSP) is a triple $\langle V,D,C \rangle$ such that $V = \{V_1, \ldots, V_k \}$ is a set of \emph{variables}, each element $D_i$ of $D = \{ D_1, \ldots, D_k \}$ is a set, called the \emph{domain} of the variable $V_i$, and each element $C_i$ of $C = \{ C_1, \ldots, C_l \}$, called a \emph{constraint}, is a subset of the Cartesian product $\bigtimes_{i=1}^k D_i$. 
An element of $\bigtimes_{i=1}^k D_i$ is called an \emph{assignment} of the CSP, and an assignment is called a \emph{solution} if it is in the intersection $\bigcap_{i=1}^l C_i$.

A \emph{permutation} of a set $\Omega$ is a bijective function from $\Omega$ to itself. 
We write permutations using the cycle notation. 
A \emph{permutation group} $G$ over $\Omega$ is a set of permutations of $\Omega$  that is closed under inverse (for all $g \in G$, its inverse $g^{-1}$ is also in $G$), closed under composition (for all $g,h \in G$, its composition $gh$ is also in $G$), and contains the identity permutation $1_G$ (the permutation that fixes all points in $\Omega$). 
The \emph{symmetric group} $\Sym(\Omega)$ of $\Omega$ is the permutation group consisting of all permutations of $\Omega$. 
A \emph{symmetry group} of a set $X$ is a group $G$ that \emph{acts} on $X$. That is, the identity $1_G$ fixes each $x \in X$ and $(x^g)^h = x^{(gh)}$ for all $x \in X$ and $g,h \in G$, where $x^g$ denotes the image of $x$ under the symmetry $g$. 
Group action formalises the notion of symmetries, and its conditions establish the expected behaviour of symmetries.

In this paper, we define a symmetry group $G$ of a CSP as a group acting on all possible assignments that preserves the solution set. This gives rise to an equivalence relation for all possible assignments (and hence the set of all solutions), where two assignments $a, a'$ are equivalent if there exists $g \in G$ such that $a^g = a'$. 
A sound \emph{symmetry-breaking constraint} is a constraint that excludes some but not all equivalent assignments. 
A sound symmetry-breaking constraint is \emph{complete} if exactly one assignment per equivalence class is retained, and is said to be \emph{incomplete} otherwise.

The \emph{lex-leader constraint} is a general symmetry-breaking constraint that asserts that only the smallest assignment of each equivalence class should be considered. For a symmetry group $G$ of the domain $\mathrm{Dom}(X)$ of a variable $X$, the full lex-leader constraint $LL_{\leq}(G, X)$ with respect to a total ordering $\leq$ of $\Dom(X)$ is the constraint $\forall g \in G. \, X \leq X^g$. Note that a symmetry-breaking constraint is sound if it is implied by the full lex-leader constraint, and is complete if the converse implication also holds.
A commonly used order is the lexicographical ordering: for a total order $\leq$ on set $S$, the \emph{lexicographical ordering $\leq_{lex}$ over $\leq$} is the total ordering on the tuples over $S$ such that $(s_1, \ldots, s_k) \leq_{lex} (s'_1, \ldots, s'_k)$ if and only if either $s_i = s'_i$ for all $i$, or there is an $i$ such that $s_i < s'_i$ and $s_j = s'_j$ for all $j < i$.

\subparagraph*{High-level modelling with Essence}
\essence is a constraint modelling language that allows the \emph{abstract type constructors} of: sets, multisets, sequences, functions, relations and partitions, as well as other \emph{concrete type constructors} such as tuples, matrices, integers and Booleans. 
A type in \essence can be an arbitrarily nested construction of these constructors. For example, we can have a multiset of partitions of tuples of Booleans. 
We say that a type is \emph{abstract} if its construction involves at least one abstract constructor, and is \emph{concrete} otherwise. 
The method proposed in this paper will only affect the symmetry breaking where we have variables of abstract type.
We use the notation \emph{matrix indexed by $[I_1, \ldots, I_k]$} to refer to a $k$-dimensional ($k \geq 1$) matrix where the values are $m[i_1] \ldots [i_k]$, for each $i_j$ in $I_j$.

\begin{remark} \label{rem: semantics of types}
We shall use \cite[Remark 1]{unnamedTypeSB} (reproduced in \Cref{def: types summary}) to express the values of each \essence type in terms of multisets, tuples and matrices. 
For example, functions $f: T_1 \rightarrow T_2$ can be seen as a set of tuples $\{(t, f(t)) \mid t \in T_1 \}$. 
Note that this gives the semantics of types which we use in the proofs, but does not necessarily give how the values are represented in the solvers. 
\end{remark}

\conjure is an automatic model rewriting tool that rewrites or \emph{refines} models in \essence into models in \essencePrime, a constraint modelling language which does not allow abstract types. 
It does so by rewriting abstract variables into their representations (a.k.a.\ viewpoints) as concrete types, possibly by imposing additional constraints (called \emph{structural constraints}), and replacing existing constraints relating to these abstract variables with a set of equivalent constraints over concrete types.

Representations are well-studied, so we will not detail them here. 
However, we will be discussing two fundamental representations of multisets (a multiset is a generalisation of sets where element repetitions are allowed): for a multiset $s$ over $T$, the \emph{explicit representation} of $s$ is an (ascending) ordered matrix whose elements are elements of $s$; the \emph{occurrence representation} of $s$ is a matrix indexed by $T$ of integers, where the entry at index $t$ is the number of occurrences of $t$ in $s$. The explicit and occurrence representations of sets are similar, but we require the matrix in the explicit representation to be strictly ascending and we have Boolean elements instead of integer elements in the occurrence case.

\begin{remark} \label{rem: representation}
It is important to note that representations are carefully designed not to introduce any symmetries, which gives us some useful functions. 
When we represent a variable $X$ of type $T$ into variable $X'$ of type $T'$,
we have an injection $\psi: \Dom(X) \rightarrow \Dom(X')$ together with a structural constraint $S$ on $X'$ such that $t' \in \mathrm{Im}(\psi)$ if and only if $t'$ satisfies $S$.
Taking the values $\Val(T)=\Dom(X)$ and $\Val(T')$ to be the values in $\Dom(X')$ that satisfy its structural constraints, we have a bijection $\phi: \Val(T) \rightarrow \Val(T')$. 
\end{remark}

For example, when we represent sets using the explicit representation, \linebreak $\Dom(X)$ and $\Val(T)$ consist of sets, $\Dom(X')$ consists of lists, and $\Val(T')$ consists of elements of $\Dom(X')$ which are strictly ordered lists.
More information on the representations and refinements used in \conjure can be found in \cite{akgun2022conjure}.

\subparagraph*{Breaking the symmetries of indistinguishable objects} A commonly occurring source of symmetries is indistinguishable objects. These are objects where permuting the objects gives an equivalent solution. These occur in many problems: e.g.\  in a delivery problem, we may treat trucks and drivers as indistinguishable; when setting up a rota for a whole work week, we may also treat days of the week as indistinguishable.
\essence allows one to express the notion of indistinguishable objects using \emph{unnamed types}, which allows correct symmetry-breaking constraints to be \emph{automatically} generated without requiring expert knowledge.

An unnamed type $T$ can be thought of as an enumerated type together with a symmetry group that allows all permutations of its values. 
(Multiple) unnamed types can be used in nested constructions of types, e.g. \texttt{matrix indexed by [T,U] of set (size 2) of T}, where $T$ and $U$ are distinct unnamed types. 
More information about unnamed types can be found in \lastPaper.
It it important to note that unnamed type symmetries do not fall neatly into well-studied symmetry classes such as variable and value symmetries, and go beyond those investigated in the literature such as \cite{WalshGeneralSBConstraints,precedeChains}. 

\lastPaper shows how the symmetries of unnamed types can be consistently broken by adding constraints of the form $X \leq X^g$ where $X$ is a variable, $g$ is a symmetry of unnamed types,  $X^g$ represents the variable after applying the induced symmetry of $g$ and $\leq$ is an ordering of the possible values of $X$.
This required defining, for each possible (nested) type, (1) the symmetry of its values induced by the symmetries of unnamed types (in Def 2), and (2) a total order for its values (in Def 4 \& Rem 2). Recall \Cref{rem: semantics of types}. We reproduce slightly generalised versions here for completeness.

\begin{definition} \label{def: induced symmetry} 
Let $G$ be a symmetry group of the values of a type $T$, and $U$ be any type (whose construction may contain $T$). Then the \emph{naturally induced action of $G$} on $\Val(U)$ is the action defined by: for all $g \in \Sym(T)$ and $u \in \Val(U)$, 
(a) if $T=U$, then $u^g$ is the image under the action on $\Val(T)$; 
(b) if $u$ not constructed from $T$, then $u^g=u$;
(c) if $u$ is a matrix indexed by $[I_1, \ldots, I_k]$ of $E$, then the image $u^g$ is a matrix where its $i$-th element $u^g[i]$ is $(u[i^{(g^{-1})}])^{g}$, where $i^{(g^{-1})}$ denotes the preimage of $i$ under $g$; 
(d) if $u$ is a multiset $\{ u_1,  \ldots, u_k\}$, then $u^g = \{u_1^g,  \ldots, u_k^g\}$; 
(e) if $u$ is a tuple $ (u_1, \ldots, u_k)$, then $u^g = (u_1^g, \ldots, u_k^g)$.
\end{definition}

\begin{definition}\label{def: ordering}
We define a total ordering $\symleq_T$ for values $\Val(T)$ of a type $T$ recursively by:
(a) if $\Val(T)$ consists of integers, we take $\symleq_T$ to be $\leq$ on integers;
(b) if $\Val(T)$ consists of Boolean, we use $\texttt{false} \symless_T \texttt{true}$;
(c) if $\Val(T)$ consists of enumerated types, then $x \symless_T y$ if $x$ occurs before $y$ in the definition of the enumerated type; 
(d) if $\Val(T)$ consists of matrices or tuples of an inner type $S$, then take $\symleq_T$ to be lexicographical order $\symleq_{S{lex}}$ over an order $\symleq_S$ for the inner type; 
(e) if $\Val(T)$ consists of multisets of type $S$, take $\symleq_T$ to be such that $x \symleq_T y$ if and only if $[-\mathrm{freq}(x,s_i) \mid 1 \leq i \leq |S|]  \symleq_{Slex} [-\mathrm{freq}(y,s_i) \mid  1 \leq i \leq |S|]$, where $\mathrm{freq}(z,a)$ denote the frequency of $a$ in set $z$, the ordering $\symleq_{Slex}$ is the lexicographical order over a total order $\symleq_S$ of $S$, and each $s_i$ is the $i$-the smallest value in $S$ w.r.t.\ $\symleq_S$. 
\end{definition}

\conjure rewrites a model with unnamed types into an \essencePrime model using these definitions of symmetries and ordering, in conjunction with the existing rewriting methods. 
The method is illustrated in the following example.

\begin{example}\label{example: unnamed}
Suppose that we have a variable $X$ of type \texttt{set (size 3)} of unnamed type $T$ of size 4.
We consider the values $\Val(T)$ of $T$ as `tagged integers' $1_T, 2_T, 3_T, 4_T$. We have put a label on the values and hence distinguished them, so any permutation of $1_T, 2_T, 3_T, 4_T$ is a symmetry of the values $\Val(T)$. 

These symmetries of $\Val(T)$ induce symmetries on the values of $X$. 
For example, consider the symmetry  $g:=(1_T, 2_T)(3_T, 4_T)$, which swaps $1_T$ and $2_T$, and simultaneously swaps $3_T$ and $4_T$. 
The image under the symmetry of a possible value $x = \{1_T, 2_T, 3_T\}$ of $X$ is $x^g:= \{1_T^g, 2_T^g, 3_T^g\} = \{2_T, 1_T, 4_T\}$, which is $\{1_T, 2_T, 4_T\}$, as order does not matter in sets. 

We add symmetry-breaking constraints for $X$ consisting of constraints of the form $X \leq X^s$ for some total ordering $\leq$ of the domain of $X$ and symmetry $s$. 
To check if the constraint is falsified for the value $x$ of $X$ and symmetry $g$, we need to check if $\{1_T, 2_T, 3_T\} \leq \{1_T, 2_T, 4_T\}$. 
Since $[-\mathrm{freq}(x, i_T) \mid 1 \leq i \leq 4] = [-1,-1,-1, 0] \symleq_{lex} [-1,-1,0,-1] = [-\mathrm{freq}(x^g, i_T) \mid 1 \leq i \leq 4]$, the constraint $X \leq X^s$ is satisfied.
\end{example}

Finally, we present a complication when breaking the symmetries of nested abstract objects. In the rest of this paper, we will more formally discuss the issues raised in \Cref{ex:introsort}, and how we solve them.

\begin{example}\label{ex:introsort}
Extending \Cref{example: unnamed}, consider a variable $Y$ of type \texttt{set (size 2) of set (size 3) of $T$}, an assignment $v:=\{ \{2_T,3_T \},\{2_T, 4_T \},\{1_T, 3_T \} \}$ of $Y$ and the permutation $g:= (1_T, 2_T)$ that swaps the values $1_T$ and $2_T$. 
Let us assume that we represent $Y$ using the explicit representation for the outer set, and the occurrence representation for the inner sets.

So $v$ is represented as $[[0,1,0,1],[0,1,1,0],[1,0,1,0]]$. Its image $v^g$ is \linebreak $\{ \{1_T,3_T \}, $ $\{1_T, 4_T \}, $ $\{2_T, 3_T \} \}$, which is represented by $[[0,1,1,0],$ $[1,0,0,1],$ $[1,0,1,0]]$. 
Note that, as $g$ maps the smaller set in $v$ to the larger set in $v^g$ and explicit representations require their elements to be ordered, we have to re-order the inner sets. 
This re-ordering of the inner sets is what makes implementing $Y \leq Y^g$ difficult. 
This constraint is currently implemented by splitting it into two constraints, $Y \leq Z$ and $Z = Y^g$. 
The first $Y \leq Z$ can be implemented as a single lexicographic ordering constraint for any type in \essence{}, but $Z=Y^g$ is more difficult. In practice, we impose constraints of the form ``for every $y \in Y$, there exists $z \in Z$, such that $y^g=z$''. These constraints are required to correctly find the image of $Y$ under $g$, but greatly increase the size of our model, slowing down the solving process.

\end{example}

\section{Representation-dependent ordering}
\label{section: dynamic ordering}

The lex-leader method uses the conjunction of constraints of the form $X \leq X^g$, where $\leq$ is a total order of $\Dom(X)$. 
The lex-leader method correctly breaks all symmetries as long as we have a total order, and does not restrict the ordering to be used. 
When using pre-determined ordering of abstract objects from the literature (such as in \cite{multisetOrdering,unnamedTypeSB}), determining if a nested set is smaller than the other can be slow.
This is because these orderings of multisets are essentially the occurrence representations of multisets. If we choose to represent multisets in another way, e.g., as explicit ordered lists, then we effectively have two representations (one for solving and one for ordering), which may not be what we want. 
Further, if the multisets are deeply nested, the representation may be very long.  

\begin{example} \label{ex: bibd running example}
Consider the Balanced Incomplete Block Design problem~\cite{csplib:prob028}), where one must find an arrangement of $v$ distinct objects into $b$ blocks of size $k$ satisfying certain constraints.  
We may model this using a decision variable \texttt{bibd} of type \texttt{set (size b) of set (size k) of T} where \texttt{T} is an unnamed type of size $v$.    
To break the symmetries of the unnamed type \texttt{T}, we add constraints of the form $\texttt{bibd} \symleq_N \texttt{bibd}^g$ for $g \in \Sym(\Val(T))$, where $\symleq_N$ is an ordering of nested sets. 

As a toy example, suppose that $b=3, k=2$ and $v=4$.  
Now, to determine if $\texttt{bibd} \symleq_N \texttt{bibd}^g$ for some symmetry $g$, if we use the method from \lastPaper, we need the frequency tuple $[-\mathrm{freq}(v, \omega_i) \mid 1 \leq i \leq |\Omega| ]$, where $\Omega$ is the set consisting of all 2-sets of $\Val(T)$. 
Since elements of $\Omega$ are also sets, we again use the frequency tuples of these inner sets to give an ordering of $\Omega$. This ordering has $\{1,2\} \symleq \{1,3\} \symleq \{1,4\} \symleq \{2,3\} \symleq \{2,4\} \symleq \{3,4\}$. 
For example, the frequency tuple of $v := \{ \{2_T,3_T \},\{2_T, 4_T \},\{1_T, 3_T \}  \}$ is $[0,-1,0,-1,-1,0]$. 

However, abstract objects would need to be represented to be fed into solvers. 
Suppose that we represent the variable \texttt{bibd} as an explicit representation of occurrence representation. This means that \texttt{bibd} is replaced with a variable \texttt{m}, of type  \texttt{matrix indexed by [int(1..b), T] of Bool}, a 2-dimensional matrix where each row of \texttt{m} represents one inner set $s_i$ of \texttt{bibd}, and $\texttt{m}[i][j]$ is true if $s_i$ contains \texttt{j}.
Then, to determine the first element of the frequency tuple of $ \texttt{bibd}$, we need to know if $\{1_T,2_T\}$ is an element of $ \texttt{bibd}$, which means that we need to know if $\bigvee_{1 \leq i \leq 3} (m[i][1_T]\wedge m[i][2_T])$. This means that in the worst case, we need to know half of the values of $m$ before we can use the lex-leader to break symmetries. 

Suppose further that we have a symmetry $g:=(1_T, 3_T)(2_T, 4_T)$, and we want to check if $ \texttt{bibd} \symleq_N  \texttt{bibd}^g$. 
Then, to determine the first element of the frequency tuple of $\texttt{bibd}^g$, again we need to know if $\{1_T,2_T\} \in \texttt{bibd}^g$, but this is only true if $\{3_T,4_T\} \in \texttt{bibd}$. Similarly to before, to determine this last inclusion, we need to know if $\bigvee_{1 \leq i \leq 3} (m[i][3_T]\wedge m[i][4_T])$. So, in the worst case, we need to know the other half of $m$ to determine the first value of the frequency tuple of $\texttt{bibd}^g$, and so we need to have all elements of $m$ to determine if $\texttt{bibd} \symleq_N \texttt{bibd}^g$. 
\end{example}

In this paper, we introduce an alternative ordering of abstract types that depends on the representation used, thereby giving us a representation-dependent ordering that is dynamic during compilation, but not during search. 
We now assume that we have constraints of the form $X \symleq_{{T}} Y$ where both sides of the inequality are of \emph{abstract} type $T$. 
These constraints would need to eventually be rewritten into equivalent constraints in terms of concrete types only, according to a choice of representation. We assume that the choice of representation for $X$ and $Y$ is the \emph{same}. This is the case when we want to break the symmetries of unnamed types, as $Y$ is always the image $X^g$ of $X$ under some symmetry $g$. 

We would want to rewrite the constraint $X \symleq_{{T}} Y$ into an equivalent constraint $X' \symleq_{T'} Y'$ for some ordering $\symleq_{T'}$ over values of $T'$, where $X$ and $Y$ are represented by expressions $X'$ and $Y'$ respectively. 
For these to be equivalent, for any $x,y \in \Dom(X)$, we need $x \symleq_{{T}} y$ exactly when $x' \symleq_{T'} y'$, where $x', y' \in \Dom(X')$ are the corresponding values representing $x$ and $y$ respectively. 
Here, we shall define a total ordering $\symleq_{{T}}$ for $\Dom({X})$ as \textit{exactly} the ordering that satisfies this requirement. Recall \Cref{rem: representation}.

\begin{definition} \label{def: dot less}
For a variable $X$ of abstract type $T$, we define a total order $\dotleq_T$ of $\Dom(X)$ by: if $X$ is to be represented with a variable $X'$ of type $T'$ via an  injection $\psi: \Dom(X) \rightarrow \Dom(X')$, then, for $a,b \in \Dom(X)$, take $a \dotleq_T b$ if and only if $\psi(a) \dotleq_{T'} \psi(b)$, where $\dotleq_{T'}$ is a total order of $\Dom(X')$.
\end{definition}

The fact that $\dotleq_T$ is well-defined since the domain of an injective function $f$ induces a total ordering of its image. The full proof can be found in \Cref{appendix: pf order well defined}.

As long as we have defined total ordering for concrete types, we obtain total orderings on all types using \Cref{def: dot less}
It is crucial that all representations of abstract variables do not introduce any symmetries, which is indeed true for \conjure.

\begin{example}
Let variable $X$ be of type sets of size 3 over $\{1, \ldots, 5\}$. 
Suppose that we take the explicit representation of sets, so a value of $X$ is represented as an ordered matrix over $\{1, \ldots, 5\}$. 
Then the ordering $\dotleq$ over $\Dom(X)$ is defined as the ordering satisfying: $\{x_1, x_2, x_3\} \dotleq \{y_1, y_2, y_3 \}$ whenever $[x_1, x_2,x_3] \leq_{lex} [y_1, y_2,y_3]$, assuming $x_1 \leq x_2 \leq x_3$ and $y_1 \leq y_2 \leq y_3$.
In particular, the set $\{1,2,3\}$ is less than the set $\{1,3,4\}$ because $[1,2,3] \leq_{lex} [1,3,4]$. 

To see why it is important that rewriting does not introduce any symmetries, suppose instead that we try to represent sets as unordered matrices. 
Then $[2,1,3]$ and $[1,2,3]$ both represent the set $\{1,2,3\}$. 
Now $[1,2,4] \leq_{lex} [2,1,3]$ gives $\{1,2,4\} \dotleq \{1,2,3\}$, 
but $[1,2,3] \leq_{lex} [1,2,4]$ gives $\{1,2,3\} \dotleq \{1,2,4\}$, a contradiction. 
\end{example}

\section{Delayed symmetry application}
\label{sec: fast sym break}

Now that the ordering is in terms of the representation used, it can still be difficult to establish if an abstract object is smaller than the other:

\begin{example}\label{ex: bibd running example 2}
Recall \Cref{ex: bibd running example}. 
As before, we add lex-leader constraints of the form $\texttt{bibd} \texttt{<=} \texttt{bibd}^g$, but this time, we have $\texttt{<=}$ to denote the representation-dependent total ordering of the possible values of $ \texttt{bibd}$. 

When \conjure rewrites $\texttt{bibd}$ into its representation $\texttt{m}$, we introduced new symmetries of \texttt{m} -- the symmetry on the first index. Typically, such a symmetry is always broken by adding constraints to order the elements of \texttt{m}.
This is indeed what is automatically done by \conjure upon the introduction of \texttt{m}. 
This ensures that we have an underlying bijection from possible values of $\texttt{bibd}$ to the possible values of $\texttt{m}$ which satisfies this additional ordering constraint. 

Let us first work out, theoretically, if a value $v:= $ $\{ \{2_T,3_T \},$ $\{2_T, 4_T \},$ $\{1_T, 3_T \} \}$ of \texttt{bibd} satisfies $\texttt{bibd} $ $\texttt{<=} $ $\texttt{bibd}^g$ for $g:= (1_T, 2_T)$. 
As seen in \Cref{ex:introsort}, the representation $v'$ of $v$ is $[[0,1,0,1],$ $ [0,1,1,0], $ $[1,0,1,0]]$, which is lexicographically smaller than the representation $(v^g)' =  [[0,1,1,0],$ $[1,0,0,1],$ $[1,0,1,0]]$ of $v^g$, so the constraint is satisfied.

When solving the problem using a constraint solver, we have to work with the lower-level Boolean matrices. 
So applying the symmetry $g$ to $\texttt{bibd}$ corresponds to swapping the first two columns of $\texttt{m}$. However, we now need to reorder the rows of \texttt{m} so that its elements are in increasing order. In our example, permuting the first two columns of $v'$ gives $v'^g:= [[1,0,0,1],[1,0,1,0],[0,1,1,0]]$, and reordering gives exactly $(v^g)'$. 
This means that determining if $\texttt{bibd} \texttt{<=} \texttt{bibd}^g$ by considering the representative values is quite difficult: we need to apply the symmetry to the representative value and then reorder. In \conjure{}, we implement this by introducing an auxiliary copy $\texttt{ma}$ of $\texttt{m}$ and constraints that enforce $\texttt{ma}$ is a correct refinement of $\texttt{bibd}^g$, where $g$ acts as described above -- including reordering the rows. This creates a significantly larger model, in terms of both variables and constraints.

Instead, we propose not to reorder after applying the symmetries to the representative value. 
We will show that not sorting the rows after applying $g$ to each row of \texttt{m} creates a sound but incomplete symmetry-breaking constraint.  This is equivalent to applying the permutation to the representation $v'$ of the set $v$ to obtain $v'^g$, instead of applying the permutation to the set itself before taking its representation $(v^g)'$, effectively delaying the symmetry application until after taking representations. 
Notice that $(v^g)'$, which is sorted, cannot be greater than $v'^g$. This means that each ordering constraint \texttt{m <= m$^g$} is implied by the ordering constraint \texttt{bibd <=  bibd$^g$}. So the constraint $LL_{\leq}(\texttt{m}, G)$, being implied by the full lex-leader $LL_{\leq}(\texttt{bibd}, G)$, is a sound symmetry breaking constraint.
\end{example}

We formalise this observation in \Cref{lemma: quick works}, whose proof is a straightforward application of the assumptions. 

\begin{lemma} \label{lemma: quick works}
Recalling \Cref{rem: representation}, suppose that a variable $X$ of type $T$ is to be represented as a variable $X'$ of type $T'$ using an underlying bijection $\phi: \Val(T) \rightarrow \Val(T')$. 
Let $\symleq_T$ and $\symleq_{T'}$ be total orderings of $\Dom(X)$ and $\Dom(X')$ respectively, such that
$x \symleq_T y \Leftrightarrow \phi(x) \symleq_{T'} \phi(y)$.  
Let $G$ be a symmetry group of the values of a type $U$ and consider the naturally induced actions of $G$. 
If $\phi$ satisfies:  
\begin{equation} \label{eqn: condition for delay}
    \phi(x^g) \symleq_{T'} \phi(x)^g \text{  for all  }g \in G \text{ and } x \in \Val(T),
\end{equation}
then for any $g \in G$, the constraint $X' \symleq_{T'} (X')^g$ is implied by the constraint $X \symleq_T X^g$. 
Further, if equality is attained in  \Cref{eqn: condition for delay}, then the two constraints are equivalent.
\end{lemma}

The condition $x \symleq_T y \Leftrightarrow \phi(x) \symleq_{T'} \phi(y)$ holds for orderings from \Cref{def: dot less}. 
If \Cref{eqn: condition for delay} also holds, then the constraint $X' \symleq_{T'} (X')^g$ can replace the constraint $X \symleq_T X^g$ as a possibly weaker symmetry-breaking constraint. Note that no constraints are added or removed in this replacement.

\Cref{lemma: quick works for occ and exp of sets} shows that \Cref{eqn: condition for delay} is satisfied for the two common representations of multisets.
A slight generalisation of the example shows that the condition holds for the explicit representation. For the occurrence representation, we have the stronger property of $\phi(x^g) = \phi(x)^g$, which follows from \Cref{def: induced symmetry}. 
See \Cref{appendix: pf of equick for sets} for the full proof.

\begin{lemma}\label{lemma: quick works for occ and exp of sets}
Let $G$ be a symmetry group of the values $\Val(U)$ of a type $U$.
For a multiset type $T$, considering the naturally induced actions of $G$, and letting $\symleq_{S}$ be a total order for the inner type.
\begin{enumerate}
    \item if we represent $T$ using the explicit representation where the elements are ordered using  $\symleq_{S}$. 
    taking $\symleq_{T'}$ to be the lexicographical order over $\symleq_{S}$, then $\phi(x^g) \symleq_{T'} \phi(x)^g$ for all $g \in G$ and $ x \in \Val(T)$. 
    \item if we represent $T$ using the occurrence representation  
    where we order the indices using $\symleq_{S}$,  
    then $\phi(x^g) = \phi(x)^g$ for all $g \in G$ and $ x \in \Val(T)$.  
\end{enumerate}
\end{lemma}

It is important to be careful with the ordering used, as there is a risk of incompatible orderings. 
We avoid this complication in our implementation by always using the natural, ascending order. 

\begin{example}\label{ex: incompatible ordering}
Let $x$ be the set $\{1,2\}$ and $g$ be the permutation $(1,3)$ that swaps $1$ and $3$. Suppose that we represent sets using the explicit representation.  
Then $x^g = \{2,3\}$ and so $\phi(x^g) = [2,3]$, while $\phi(x) = [1,2]$ and so $\phi(x)^g = [3,2]$. In this case $\phi(x^g) \leq_{lex} \phi(x)^g$.

Let $\leq_{\overline{lex}}$ be the lexicographical order that compares the elements in descending order. Then clearly it is not true that $\phi(x^g) \leq_{\overline{lex}} \phi(x)^g$. This means that, if we choose to order lists using this ordering instead of the usual lex-ordering, our method may not work. 
Similarly, if the refinement represents sets as descending ordered lists, then $\phi(x^g) = [3,2]$, while $\phi(x) = [2,1]$ and so $\phi(x)^g = [2,3]$. Here again it is not true that $\phi(x^g) \leq_{lex} \phi(x)^g$.
\end{example}

\section{Breaking the symmetries of unnamed types}
\label{sec: application}
\label{sec: discussion}

We apply our method to breaking the symmetries induced
by unnamed types in \essence. 
Other technologies can follow a similar method to handle  symmetries of indistinguishable objects.

\subsection{Method summary}
\label{sec: implementation}

We start with an \essence model $M$ with unnamed types $T_1, T_2, \ldots, T_k$. We obtain an \essencePrime model where the unnamed type symmetries are soundly (but possibly incompletely) broken using the following steps. 

\begin{enumerate}
    \item Replace each unnamed type $T_i$ by tagged integers of values $1_{T_i}, 2_{T_i}, \ldots, $ $ |T_i|_{T_i} $. 
    \item Letting $V_1,V_2, \ldots, V_n$ be the decision variables of $M$, let $X$ be a new decision variable that represents the tuple $(V_1, V_2,  \ldots ,V_n)$.
    \item Add constraints of the form $X \texttt{.<=} \, \texttt{transform}(gs,X)$ for some permutation combinations $gs = (g_1, g_2, \ldots, g_k)$, where each $g_i \in \Sym(\Val(T_i))$.  Here, $\texttt{.<=}$ is $\dotleq_T$ from \Cref{def: dot less} for the appropriate type $T$, and $\texttt{transform}(g,X)$ represents the image of $X$ under the symmetry $g$ from \Cref{def: induced symmetry}. 
    \item Go through \conjure refinements to rewrite abstract variables into concrete variables, according to some representation.
    \begin{enumerate}
        \item If $X$ is abstract and is refined to an expression $X'$, then the constraint $X \texttt{.<=}  \texttt{transform}(gs,X)$ is rewritten to $X' \texttt{.<=}  \texttt{transform}(gs,X')$. 
        \item If $X$ is concrete, then it is rewritten to $X$ $\texttt{<=}  $ $\texttt{transform}(gs,X)$, for an appropriate concrete ordering $\texttt{<=}$. The expressions of form $\texttt{transform}(g, X)$ are subsequently removed using the definition of symmetry application on concrete type  from \Cref{def: induced symmetry}.  
    \end{enumerate}
\end{enumerate}

Steps 1--3 follow the procedure in \cite[Proposition 1]{unnamedTypeSB} to obtain a model without unnamed types. 
However, we replace the static ordering $\symleq$ from \Cref{def: ordering} with our representation-dependent ordering $\dotleq$ from \Cref{def: dot less}. 
This is correct because lex-leader expressions require a total order, but it does not matter which.
If $T'$ is concrete, we replace $\dotleq_{T'}$ with an ordering of concrete type (Step 4b), which should natively exist in most solvers. We never need to implement an ordering for abstract types, in contrast to the methods in \lastPaper.

In Step 4(a), constraints of the form $X \dotleq_{T} Y$, are then rewritten (and possibly weakened) into $X' \dotleq_{T'} X'^g$ using \Cref{lemma: quick works}.  
The condition ``$x \symleq_T y \Leftrightarrow \phi(x) \symleq_{T'} \phi(y)$'' is attained since we use the ordering from \Cref{def: dot less}. 
However, for this method to be sound, we need to show that \emph{all} representations in \conjure satisfy \Cref{eqn: condition for delay}:

\begin{lemma} \label{lemma: quick is ok for conjure rep}
Let $G$ be a symmetry group of the values $\Val(U)$ of a type $U$.
For each representation in \conjure, letting the underlying bijection be $\phi: \Val(T) \rightarrow \Val(T')$ and considering the natural induced actions of $G$, we have $\phi(x^g) \dotleq_{T'} \phi(x)^g$ for all $x \in \Val(T)$ and $g \in G$, where $\dotleq_{T'}$ is as defined in \Cref{def: dot less}.
\end{lemma}

More details on the proof of this claim can be found in \Cref{subsec: pf rep order}.
As observed in \cite[Remark 1]{unnamedTypeSB}, all types in \conjure can be expressed in terms of multisets (which are abstract), matrices and tuples (which are concrete and hence do not have to be represented). So every representation of abstract types used in \conjure is a generalisation of the two representations of multisets from \Cref{lemma: quick works for occ and exp of sets}.

\subsection{Experiments}
\label{sec: results}

We compare the performance of our method against that in \lastPaper. All experiments are run on Core i7-13700HX, Ubuntu 25.04 with 32 GB RAM. See the code appendix to see how to reproduce the results. 

So we can focus on the performance of symmetry breaking, rather than considering complete problems, we will consider finding all assignments to a single abstract variable, which is constructed from unnamed types.
We also do not dictate which representations to use, but instead let \conjure decide the most suitable ones, and report those that use a representation where our method produces a different constraint, such as when we have explicit representations of sets.
We believe these experiments show the potential benefits of our new technique. 

We summarise the results of our experiment in \Cref{tab: expt results}. The `\textbf{Old}' and `\textbf{New}' columns denote the method in \lastPaper and our new method, respectively. We consider six different high-level variables, each built using a single unnamed type. For each problem, we select the smallest instance where the `Old' technique takes more than a second. The time taken varies between experiments because the set of solutions to each of our problems grows extremely rapidly as the parameters are increased. The exact instances used are provided in the supplementary data.

All experiments are run with the \texttt{Consecutive} and \texttt{Independently } options, the best performing options from \lastPaper. 
The \texttt{Consecutive} option generates lex-leader for only the permutations that swap consecutive values of the unnamed type. 
We do not report the timings for \texttt{AllPairs} (which uses all permutations that swap 2 values), as they show a similar pattern. 
We also do not report the timings when using all permutations (which gives the full lex-leader with `Old') because it is too slow, especially for the `Old' method, as it introduces an exponential number of symmetry-breaking constraints.
Our experiments also do not include models with more than one unnamed type, as the number of unnamed types is irrelevant to our new method. We have performed further tests with multiple unnamed types, and the results are similar to those presented here. As these experiments only include one unnamed type, the other incomplete options of \texttt{Independently} and \texttt{Altogether} from \lastPaper are equivalent. 

The problems are solved using the constraint solver \minion \cite{minion}, and \savilerow \cite{sr-journal-17} is used to convert the \essencePrime outputted by \conjure into input for \textsc{Minion}.
The 3 inner rows denote the total number of solutions, the number of nodes in \minion and the total time taken, in seconds. 

\begin{table*}[t] 
\small
\centering
\begin{tabular}{c|rr|rr|rr}
\hline \hline
 \multicolumn{1}{c|}{Instance} & \multicolumn{2}{c|}{\textsc{function}} & \multicolumn{2}{c|}{\textsc{set of function}} & \multicolumn{2}{c}{\textsc{set of matrix}} \\
 \multicolumn{1}{c|}{} & Old & New & Old & New & Old & New  \\
\hline

 Solutions& 1,469,103 & 1,492,818 & 25,612 & 34,790 & 5,621 & 9,979 \\
 Nodes& 2,982,777 & 4,512,188 & 637,244 & 70,570 & 526,917 & 19,970 \\
 Time & 27. & 15. & 7.2 & 0.07 & 3.5 & 0.01 \\

\hline 

\hline
\hline
\multicolumn{1}{c|}{Instance} & \multicolumn{2}{c|}{\textsc{set of multiset}} & \multicolumn{2}{c|}{\textsc{set of relation}} & \multicolumn{2}{c}{\textsc{set of set}} \\
\multicolumn{1}{c|}{} & Old & New & Old & New & Old & New  \\
\hline

Solutions& 18 & 21 & 5,761,575 & 8,775,909 & 1,019 & 3,897 \\
Nodes& 91,358 & 52 & 114,372,858 & 17,551,836 & 388,034 & 7,944 \\
Time& 1.70 & 0.00 & 895. & 1.78 & 3.2 & 0.00 \\

\hline

\end{tabular}
\caption{Experimental results comparing the old and the new methods for various variable types}
\label{tab: expt results}
\end{table*}

\begin{table*}
\small
\centering
\begin{tabular}{l|rr|rr|rr|rr}
\hline \hline
Parameter & \multicolumn{2}{c|}{\textsc{High New}} & \multicolumn{2}{c|}{\textsc{High Old}}  & \multicolumn{2}{c|}{\textsc{High NoSym}} & \multicolumn{2}{c}{\textsc{Matrix DoubleLex}} \\
& Nodes & Time & Nodes & Time & Nodes & Time & Nodes & Time \\
\hline
14-6-2& \(1.04e2\)& 0.00& -& -& \(1.11e8\)& \(6.67e2\)& \(1.30e3\)& 0.01\\
8-4-3& 35& 0.00& \(2.78e6\)& \(1.41e2\)& \(2.51e2\)& 0.00& \(1.02e2\)& 0.00\\
8-4-4& \(1.86e3\)& 0.00& \(1.13e5\)& \(3.97e1\)& \(1.13e5\)& 0.29& \(1.14e4\)& 0.02\\
\hline

\hline \hline
Parameter & \multicolumn{2}{c|}{\textsc{Low New}} & \multicolumn{2}{c|}{\textsc{Low Old}}  & \multicolumn{2}{c|}{\textsc{Low NoSym}} & \multicolumn{2}{c}{\textsc{Matrix NoSym}} \\
& Nodes & Time & Nodes & Time & Nodes & Time & Nodes & Time \\
\hline
14-6-2& 81& 0.00& 81& 0.01& -& -& -& -\\
8-4-3& 22& 0.00& 22& 0.00& \(6.80e2\)& 0.00& \(6.80e2\)& 0.00\\
8-4-4& \(8.40e3\)& 0.01& \(8.40e3\)& 0.01& -& -& -& -\\
\hline
\end{tabular}

\caption{Solving BIBD with a variety of symmetry breaking methods, where `-' denotes timeout after 3600 seconds
}
\label{tab:bibd}
\end{table*}

The main takeaway from our results is that while our method usually produces more solutions, the time taken is significantly shorter, often by orders of magnitude. 
This pattern continues for larger instances; we found larger instances of our problems, which could be solved in a few seconds by our new method, but timed out after a day with the old method.
Furthermore, and more surprisingly, the number of search nodes is often smaller by orders of magnitude as well. This is, however, not true in the single \textsc{function} case as it is the simplest type we consider, and so the gain of our technique is smallest here -- although it still outperforms the old method in terms of runtime.

\subsection{Case Study: BIBD}

As a complete example, we consider finding one solution for the Balanced Incomplete Block Design problem (for the definition, see \Cref{ex: bibd running example}). Given BIBD parameters $v$-$b$-$k$, we model the central BIBD variable in 3 ways:
\begin{itemize}
    \item \textbf{High}: A \texttt{set (size b) of set (size k) of Obj} for a new unnamed type \texttt{Obj} of size \texttt{v}.
    \item \textbf{Low}: A Boolean 2-D matrix indexed by two unnamed types, \texttt{Obj} of size \texttt{v} and \texttt{Blocks} of size \texttt{b}.
    \item \textbf{Matrix}: A 2-D Boolean matrix indexed by integer ranges, \texttt{1..v} and \texttt{1..b}.
\end{itemize}
We consider breaking symmetries of the first two models using \texttt{Consecutive} and \texttt{Independently} generation, with both the \textbf{Old} and our \textbf{New} symmetry-breaking method, or with no added symmetry-breaking constraints (\textbf{NoSym}). In the 3rd model, we consider using ``double-lex'' (\textbf{DoubleLex}), or no symmetry breaking (\textbf{NoSym}). 
The models can be found in the supplementary data, and the results are given in \Cref{tab:bibd}.
We see the following results: 

\begin{itemize}
    \item \textsc{Low NoSym} and \textsc{Matrix NoSym} do worst, as these are equivalent. This shows that any symmetry breaking is better than no symmetry breaking.
    \item \textsc{High NoSym} does better than other forms of no symmetry breaking because \conjure{} always breaks symmetry whenever a set is turned into a matrix -- this cannot be turned off.
    \item In the \textsc{Low} model, the \textsc{New} and \textsc{Old} symmetry breaking are identical -- this is expected, as this model contains no abstract types. 
    \item When using ``New'' symmetry breaking, \textsc{High} now produces equally good results to the other models (both \textsc{Low} models) that are equivalent to \textsc{DoubleLex}. The variations come about from variables and constraints being outputted in different orders.
\end{itemize}

These results show the efficiency of double lex -- when it is being used, all models finish in almost no time. We could look at larger experiments to see where double-lex performs worse, but this would not give any insights for this paper, and is already a very well-studied problem.

\section{Conclusion and future work}
\label{sec: conclusion}

In this paper, we show that symmetry breaking for abstract types can be quite difficult, and propose a new symmetry-breaking method using two innovations: using representation-specific ordering instead of a pre-defined ordering, and breaking symmetries using the images of \emph{representations} of abstract values instead of the images of the abstract values themselves. 
We show how the method can be combined with the methods in \lastPaper to give faster symmetry breaking of unnamed types, which represent the commonly occurring indistinguishable objects.
Our method also does not require implementation of the symmetry application and ordering of \textit{abstract} types given in \lastPaper, simplifying implementations. 
While the method is not complete, we show that it can be much faster.

This method gives another dimension of variability for the incomplete symmetry breaking for indistinguishable objects. 
Comparing these choices and deciding the level of symmetry breaking for a given problem is future work. 
Another interesting direction is to investigate whether we can use the method to find or count all solutions modulo symmetries.

\section*{Acknowledgements}

This work was supported by the International Science Partnerships Fund (ISPF) and the grant was awarded by the Engineering and Physical Sciences Research Council on behalf of UK Research and Innovation (UKRI) [EP/Y000609/1].

\bibliographystyle{plain}
\bibliography{bib}

\clearpage
\newpage

\begin{appendices}

\section{\Cref{def: dot less} is well-defined}
\label{appendix: pf order well defined}

It is an elementary property of total orderings that a total ordering of the domain of an injective function $f$ induces a total ordering of its image, by setting $f(x) \leq f(y)$ whenever $ x \leq y$. 
To see that $\dotleq_T$ is well-defined, take $f$ to be the inverse of the bijection $\overline{\psi}$ obtained by restricting the range of $\psi$ to $\mathrm{Im}(\psi)$.

\section{Proof of \Cref{lemma: quick works for occ and exp of sets}}
\label{appendix: pf of equick for sets}

(a): The explicit representation of multisets is a bijection $\phi$ which maps a multiset $x$ to a matrix $m$ such that the number of occurrences of each $i$ in $x$ is the same as the number of occurrences of $i$ in $m$, with the structural constraints that elements of $m$ is in ascending order with respect to $\symleq_{S}$. 
The elements of $\phi(x^g)$ and $\phi(x)^g$ are necessarily the same (with the multiplicities taken into account), but $\phi(x^g)$ is ordered, and so must be lexicographically at most $\phi(x)^g$. 

\noindent (b): The representation has an underlying bijection $\phi$ which maps a multiset $x$ to a matrix $m$ indexed by $U$ of integers such that $m[i]$ gives the number of occurrences of $i$ in $x$.
The $i$-th entry $\phi(x^g)$ is $k$ if and only if $i$ occurs in $x^g$ with $k$ occurrences, which means that $i^{g^{-1}}$ (the image of $i$ under the inverse of $g$) occurs $k$ times in $x$. This happens exactly when the $(i^{g^{-1}})$-th element of $\phi(x)$ is $k$.
From \Cref{def: induced symmetry}, the $i$-th element of $\phi(x)^g$ is $(\phi(x)[i^{g^{-1}}])^g$, which is $k^g=k$, as required. 

\section{Proof of \Cref{lemma: quick is ok for conjure rep}}
\label{subsec: pf rep order}

In this section, we give a sketch for the proof of \Cref{lemma: quick is ok for conjure rep}.  
A summary of the representation in \conjure can be found in \cite[Table 5 \& Section 4.1.6]{akgun2022conjure}, and we shall briefly describe those as we need them.

To reduce the amount of setup, we take $g$ as a permutation representing an unnamed type symmetry, and $x$ a value of the appropriate type. We also simplify the notation to take $\leq$ as $\dotleq_T$ from \Cref{def: dot less} for an appropriate $T$, so we always have $x \leq y$ if and only if $\phi(x) \leq \phi(y)$. 
We will be using \cite[Remark 1]{unnamedTypeSB} from \Cref{rem: semantics of types}, to express the values of each \essence type in terms of multisets, tuples and matrices.
We reproduce the full statement here for reference. 

\begin{remark}\label{def: types summary}\label{def: semantics of types} \label{remark: values of types as sets and tuples}
For a type $T$, we denote its set of all possible values by $\Val(T)$, which is defined in terms of matrices, multisets and tuples:
the values of \texttt{bool}, \texttt{int} and \texttt{enum} are what one would expect; 
the values of a \texttt{tuple}, \texttt{record}, \texttt{variant} and \texttt{sequence} can be naturally defined as tuples; the values of a \texttt{matrix} are matrices,
the values of a \texttt{set}, \texttt{mset}, \texttt{partition} can be naturally defined as (nested) multisets; for types $\tau_i$, the values of a \texttt{function $\tau_1 \rightarrow \tau_2$} are  subsets of $\Val(\tau_1) \times \Val(\tau_2)$ such that there are no two elements with the same value in its first position; 
the values of a \texttt{relation $(\tau_1 * \tau_2 * \ldots * \tau_k)$} are subsets of $\Val(\tau_1) \times \Val(\tau_2) \times  \ldots \times \Val(\tau_k)$. 
\end{remark}

Note that these give the semantics of types which we use in the proofs, but do not necessarily give how the values are represented in \conjure.

We assume that all orderings of matrices are lexicographical ordering over the most natural ascending order.
Further, as in \Cref{lemma: quick works for occ and exp of sets}, representations akin to explicit and occurrence representations of multisets will have their elements and indices ordered using this natural ascending order.
So we will not encounter the complication of incompatible ordering as in \Cref{ex: incompatible ordering}.

\subsection{Representations of sets and multisets}
\label{subsec: reps of sets}

We start with the representations of sets and multisets. The proofs for other representations will be based on those here. First note that the \textbf{\texttt{Occurrence}} and \textbf{\texttt{ExplicitRepetition}} representations of multisets are the occurrence and explicit representations discussed in \Cref{lemma: quick works for occ and exp of sets}. 

\paragraph*{\texttt{Occurrence} \& \texttt{Explicit} representations for sets} The \texttt{occurrence} representation of sets is similar to the occurrence representations of multisets, but with Booleans indicating membership instead of using integers to indicate the number of occurrences. 
The \texttt{explicit} representation of sets is also similar to the explicit representation of multisets, but with repetitions disallowed.
In both cases, the proof of $\phi(x^g) \leq \phi(x)^g$ is similar to the proof of \Cref{lemma: quick works for occ and exp of sets}. 

\paragraph*{\texttt{ExplicitVariableSizeMarker} representation for sets} We represent a set $s$ with maximum size $k$ as a matrix of size $k$. There is an accompanying integer to denote the actual size of $s$ and we use \textsc{dontCare} constraints (see \cite{akgun2014breaking} for more details) to set the values at indices more than $|s|$ to the smallest values possible. 
The prefixes of length $s$ must be sorted, so these are exactly the explicit representations of sets. 

More specifically, $\phi$ maps set $x$ of maximum size $s$ to a matrix $m$ of length $s$ where the prefix $m[1,\ldots, |x|]$ is the sorted elements of $x$, and the suffix $m[|x|+1, \ldots, s]$ is the smallest possible matrix with elements not in $x$. We know from the explicit representations of sets that $\phi(x^g)[1,\ldots,|x|] \leq \phi(x)^g[1,\ldots,|x|]$. If it is strictly less than, then $\phi(x^g) $ is strictly less than $\phi(x)^g$ as well. 
If otherwise we have an equality, since $\phi(x^g)[|x|+1, \ldots, s]$ must be smallest possible, it must be at most $\phi(x)^g[|x|+1, \ldots, s]$, and so again we have $\phi(x^g) \leq \phi(x)^g$.

\paragraph*{\texttt{ExplicitVariableSizeFlags} representation for sets} Here we also represent a set $x$ with maximum size $s$ as a matrix of length $s$, but now we have an accompanying matrix $b$ of Booleans of length $s$ to indicate which indices give the elements of the set. 
We require the \textsc{True}'s in $b$ to come before any \textsc{False}'s. We again use \textsc{dontCare} constraints, now to set the values $i$ where $b[i] = \texttt{False}$ to the smallest possible ones. 

So again we have a bijection $\phi$ which maps a set $x$ of maximum size $s$ to a matrix $m$ of length $s$ where the prefix $m[1,\ldots, |x|]$ is the sorted elements of $x$ and the suffix $m[|x|+1, \ldots, s]$ is smallest possible. 
Note that $b$ remains unchanged with the application of symmetry~$g$.
So, similar to \texttt{ExplicitVariableSizeMarker}, we have $\phi(x^g) \leq \phi(x)^g$.

\paragraph*{\texttt{ExplicitVariableSizeDummy} representation for sets} We represent sets over $T$ as matrices over $\overline{T}$, where the values of $\overline{T}$ are the values of $T$ plus a dummy value \textsc{d}. We take \textsc{d} to be the largest value in $\overline{T}$, and it is fixed by the unnamed type symmetry $g$. This dummy value is used to populate the matrices when the set size is less than the maximum size. Since the matrices have to be sorted, these \textsc{d}'s are pushed to the right of the matrix. So a set $x$ of maximum size $s$ is mapped by $\phi$ onto a sorted matrix of length $s$ where the first $|x|$ elements are the explicit representation of $x$, and is followed by $s-|x|$ number of \textsc{d}'s. 
Since \textsc{d} is fixed by $g$, it follows from the explicit representation of sets that $\phi(x^g) \leq \phi(x)^g$.

\paragraph*{\texttt{ExplicitFlags} representation for multisets} This is similar to the ExplicitVariableSizeFlags representation of sets, but with an auxiliary matrix $n$ over integers that records the number of occurrences instead of using Booleans to indicate membership. 
In matrix $n$, we would similarly require the non-zero values to come before the $0$'s. 
As in ExplicitVariableSizeFlags, this matrix $n$ is unchanged by $g$. 
The proof of $\phi(x^g) \leq \phi(x)^g$ works in the same way as for ExplicitVariableSizeFlags.

\subsection{Representations of other types}

Here we consider the representations of other types. Most will have proofs similar to or follow from those in \Cref{subsec: reps of sets}, or are trivial, so we do not describe each of them in full detail.

\paragraph*{\texttt{FunctionAsMatrix} representation for functions} Recall from \Cref{def: types summary} that values of functions from $T$ to $U$ is a subset of $T \times U$. The bijection $\phi$ maps such a set $x$ of $T$-$U$ tuples to a matrix $m$ indexed by $T$ of $U$ such that $(t,u) \in x \Leftrightarrow m[t]=u$. We shall show that $\phi(x^g) = \phi(x)^g$. 

For an arbitrary $i$, let $j := \phi(x^g)[i]$. We want to show that $\phi(x)^g[i]=j$.
From the definition of $j$, we have $(i,j) \in x^g$, which means that $(i^{g^{-1}}, j^{g^{-1}}) \in x$ and so $\phi(x)[i^{g^{-1}}] = j^{g^{-1}}$. 
Recall from \Cref{def: induced symmetry} that $\phi(x)^g[i] = (\phi(x)[i^{g^{-1}}])^g$, which is $(j^{g^{-1}})^g = j$, as required.

\paragraph*{\texttt{FunctionAsMatrixPartial} representation for functions} We generalise \linebreak \texttt{FunctionAsMatrix} to support possibly partial functions. Here we also have a bijection $\phi$ that maps a set $x$ of $T$-$U$ tuples to a matrix $m$ indexed by $T$ of $U$ such that $(t,u) \in x \Leftrightarrow m[t]=u$, but also has another matrix $b$ indexed by $T$ of Booleans as in \texttt{ExplicitVariableSizeFlags} where $b[t] = \texttt{True}$ if and only if there is a pair $(t,u)$ in $x$ (indicating that $t$ has an image under the partial function). We also use \texttt{dontCare} constraints to make $[m[i] \mid b[i]=\texttt{False}]$ to be the smallest possible. Note also that $b$ is unchanged by $g$. 

To see why $\phi(x^g) \leq \phi(x)^g$, consider the first $i$ such that $b[i] = \texttt{False}$. Then the elements in $\phi(x^g)$ preceding index $i$ is a prefix of the \texttt{FunctionAsMatrix} representation of $x$, so it is equal to $\phi(x)^g[1,\ldots,i-1]$. 
Because of the \textsc{dontCare} constraint, either $\phi(x^g)[i] < \phi(x)^g[i]$, in which case we have $\phi(x^g) < \phi(x)^g$ again, or $\phi(x^g)[i] = \phi(x)^g[i]$. If it is the latter, we have $\phi(x^g)[1, \ldots, i] = \phi(x)^g[1, \ldots, i]$. We then consider the next $j$ such that $b[j]= \texttt{False}$ and similar arguments hold for indices $i+1, \ldots, j$, until we have shown $\phi(x^g) < \phi(x)^g$ or we reach the end to have  $\phi(x^g) = \phi(x)^g$.

\paragraph*{\texttt{FunctionAsMatrixDummy} representation for functions} This similarly generalises  \texttt{FunctionAsMatrix}, but uses a dummy value as in the \linebreak \texttt{ExplicitVariableSizeDummy} representation for sets. More specifically, $\phi$ maps a set $x$ of $T$-$U$ tuples to a matrix $m$ indexed by $T$ of $\overline{U}$, where $\Val{(\overline{U})} = \Val(U) \cup \{\textsc{d} \}$. 
We show that $\phi(x^g)$ and $\phi(x)^g$ have \textsc{d} at the same indices, then it follows from the proof for \texttt{FunctionAsMatrix} that $\phi(x^g) = \phi(x)^g$. 

Let $i$ be such that $\phi(x^g)[i] = \textsc{d}$. Then it means that $x^g$ does not contain a tuple which has $i$ in its first index (meaning $i$ is undefined in the partial function represented as $x^g$), which means that $x$ does not contain a tuple which has $i^{g^{-1}}$ in its first index ($i^{g^{-1}}$ is also undefined in $x$), giving us $\phi(x)[i^{g^{-1}}] = \textsc{d}$. 
From \Cref{def: induced symmetry}, $\phi(x)^g[i] =  (\phi(x)[i^{g^{-1}}])^g$, which is $\textsc{d}$ as $g$ fixes \textsc{d}.

\paragraph*{\texttt{RelationAsMatrix} representation for relations} Recall from \Cref{def: types summary} that values of a relation of $T_1, \ldots, T_k$ is a subset of the Cartesian product $T_1 \times \ldots \times T_k$. The bijection $\phi$ maps such a set $x$ to a $k$-dimensional matrix indexed by $[T_1, \ldots, T_k]$ of Booleans such that $(t_1, \ldots, t_k) \in x \Leftrightarrow m[t_1, \ldots, t_k] = \textsc{True}$. The proof of this is very similar to that of the \texttt{Occurrence} representation for multisets, but generalised to higher-dimensional matrices. 

\paragraph*{\texttt{Occurrence} representation for partitions} A partition $P$ of integers is first considered as an ordered partition $P'$, where the ordering is based on the smallest element of each cell of the partition. Then this ordered partition is represented as a matrix over integers $1, \ldots, |P|$ where $m[i]=k$, which means that $i$ is in the $k$-th cell of $P'$. Take $\phi$ to be the bijection that maps such $P$ to such $m$. 
Since we order the partition based on the smallest element of each cell, $\phi(x)$ is the smallest possible among its equivalences if we allow the relabelling of the cell numbers. 

We show that $\phi(x^g)$ and $\phi(x)^g$ have to be equivalent in this relabelling as well, from which $\phi(x^g) \leq \phi(x)^g$ will follow. 
To see that $\phi(x^g)$ and $\phi(x)^g$ are equivalent up to relabelling, notice that
$\phi(x^g)[i] = \phi(x^g)[j]$ if and only if $i$ and $j$ are in the same cell of $x^g$, which means that $i^{g^{-1}}$ and $ j^{g^{-1}}$ are in the same cell of $x$, and so $\phi(x)[i^{g^{-1}}] = \phi(x)[j^{g^{-1}}]$. Using \Cref{def: induced symmetry} again gives $\phi(x)^g[i] = \phi(x)^g[j]$.

\paragraph*{\texttt{ExplicitBounded} representation for sequences} 
Similar to \linebreak \texttt{ExplicitVariableSizeMarker}, we represent a sequence with maximum length $l$ using a matrix of length $l$ along with an integer that stores the length of the sequence. 
\Cref{def: types summary} defines values of sequences as tuples. So we have a bijection $\phi$ which maps a tuple $x = (t_1, \ldots, t_k)$ to a matrix $m$ of length $l$ where 
$m[i] = t_i$ for all $i \leq k$. We also use \textsc{dontCare} constraints to make the values of the suffix $m[k+1, \ldots, l]$ as small as possible.   
Clearly the prefixes $\phi(x^g)[1, \ldots, k]$ and $\phi(x)^g[1, \ldots, k]$ are equal. Since $\phi(x^g)[k+1, \ldots, l]$ is smallest possible, $\phi(x^g) \leq \phi(x)^g$.

\paragraph*{Other representations} Functions, relations and partitions are defined in terms of sets of tuples in \Cref{def: types summary}. So the \textbf{\texttt{FunctionAsRelation}}, \linebreak\textbf{\texttt{RelationAsSet}} and \textbf{\texttt{PartitionAsSet}} representations for functions, relations and partitions, respectively, each have the identity map as $\phi$.

\end{appendices}

\end{document}